\documentclass[journal]{IEEEtran}
\usepackage[T1]{fontenc}
\usepackage{graphicx}
\usepackage{times}
\usepackage{helvet}
\usepackage{courier}
\usepackage{amsmath}
\usepackage{algorithm}
\usepackage{algorithmic}
\usepackage{csquotes} 
\usepackage{color}
\usepackage{paralist}
\usepackage{amssymb}
\usepackage{indentfirst}
\usepackage{subfigure}
\usepackage{float}
\usepackage{multirow}
\usepackage{cite}
\usepackage{mathrsfs}

\usepackage{mathrsfs} 
\usepackage{amsfonts}
\usepackage{todonotes}
\usepackage{pgfplots} 
\usepackage{epstopdf}
\usepackage{epsfig}
\pgfplotsset{compat=newest}
\usepackage{color}
\definecolor{forestgreen}{RGB}{0,139,69}

\usepackage{xcolor}
\definecolor{citecolor}{HTML}{0071bc}
\usepackage[colorlinks, linkcolor=red,  anchorcolor=blue, citecolor=citecolor]{hyperref} 

\usepackage{xcolor}
\definecolor{SeaGreen4}{RGB}{0,205,102} 
\definecolor{SlateBlue}{RGB}{106,90,205} 
\definecolor{DarkRed}{RGB}{178,34,34} 
	
\usepackage[switch]{lineno}

\usepackage{textcomp,booktabs}
\usepackage{amssymb}
\usepackage{pifont}

\usepackage{makecell}

\usepackage{colortbl}
\definecolor{mygray}{gray}{.9}
\definecolor{mypink}{rgb}{.99,.91,.95}
\definecolor{mycyan}{cmyk}{.3,0,0,0}

\begin{document}

\title{ 
    Dynamic Parsing and Updating Natural Language Specification using VLMs for Robust Vision-Language Tracking 
}

\author{Xiao Wang, \emph{Member, IEEE}, Liye Jin, Dan Xu, Yuehang Li, Lan Chen*,  
    Yaowei Wang, \emph{Member, IEEE}, \\ Yonghong Tian, \emph{Fellow, IEEE}, Jin Tang 
\thanks{$\bullet$ Xiao Wang, Liye Jin, Dan Xu, Yuehang Li, and Jin Tang are with the School of Computer Science and Technology, Anhui University, Hefei 230601, China. (email: \{xiaowang, tangjin\}@ahu.edu.cn, jinliye@stu.ahu.edu.cn, 18856245162@163.com, liyuehang@stu.ahu.edu.cn)} 
\thanks{$\bullet$ Lan Chen is with the School of Electronic and Information Engineering, Anhui University, Hefei 230601, China. (email: chenlan@ahu.edu.cn)} 
\thanks{$\bullet$ Yaowei Wang is with Harbin Institute of Technology, Shenzhen, China; Peng Cheng Laboratory, Shenzhen, China. (email: wangyw@pcl.ac.cn)} 
\thanks{$\bullet$ Yonghong Tian is with Peng Cheng Laboratory, Shenzhen, China; School of Computer Science, Peking University, China; Shenzhen Graduate School, Peking University, China. (email: yhtian@pku.edu.cn)} 
\thanks{* Corresponding Author: Lan Chen} 
}

\markboth{ IEEE Transactions on ***, 2026 } 
{Shell \MakeLowercase{\textit{et al.}}: Bare Demo of IEEEtran.cls for IEEE Journals}

\maketitle

\begin{abstract} 
Vision-language tracking guided by natural language specifications leverages high-level semantic cues of target objects to substantially boost tracking accuracy and robustness. Existing studies have verified that adaptively optimizing textual descriptions throughout the tracking process can effectively mitigate the semantic-visual mismatch induced by dynamic variations in target appearance, position, and other inherent attributes. Nevertheless, mainstream methods that directly generate textual information via sequence models or large language models inevitably suffer from inherent defects, including erroneous target updating, excessive background distraction, and pervasive hallucination artifacts. To address the aforementioned limitations, this paper proposes a novel language dependency parsing mechanism to precisely distill core tracking principal components, encompassing target objects, semantic concepts, and background contextual information. On this basis, we perform component-aware adaptive textual description updates by exploiting the powerful cross-modal understanding capability of the pre-trained vision-language model Qwen-VL. By integrating the proposed elaborately-designed modules into the baseline framework, our method achieves consistent and superior tracking performance on multiple large-scale vision-language tracking benchmarks, including TNL2K, LaSOT, TNLLT, and OTB-LANG. The source code and pre-trained models will be released at \url{https://github.com/Event-AHU/Open_VLTrack}.
\end{abstract}


\IEEEpeerreviewmaketitle

\section{Introduction}

\IEEEPARstart{T}{racking} by natural language specification~\cite{li2017tracking, wang2021tnl2k} aims to localize and predict the trajectory of arbitrary targets in video sequences according to given linguistic descriptions, which leverages high-level semantic information to establish robust cross-modal matching between language cues and visual contents and thus achieves accurate and reliable target tracking in complex and dynamically changing scenarios. It offers significant value for research and practical deployment across UAV tracking, autonomous driving, and human-computer interaction.

\begin{figure*}
\centering
\includegraphics[width=\linewidth]{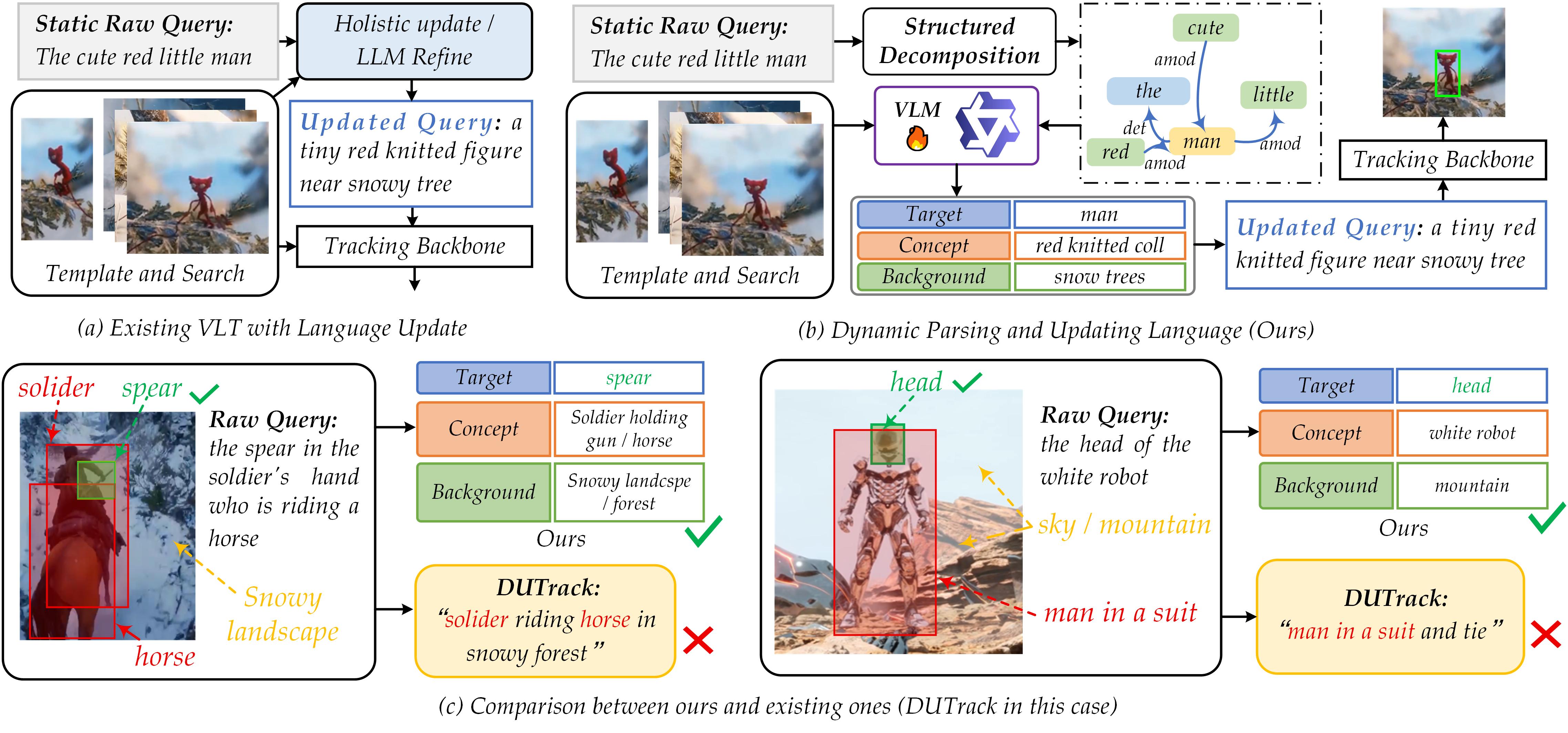}
\caption{\textbf{Comparison between existing language-updating VLT paradigms and our proposed fine-grained text update framework.}
(a) Existing VLT methods typically perform holistic language updating, where the updated description may drift toward salient objects, background context, or hallucinated cues. (b) Our framework introduces a fine-grained text update paradigm by decomposing the query into target, concept, and background fields, keeping the target identity stable, and selectively refining adaptive components according to visual changes. (c) Two representative cases illustrate the resulting difference.} 
\label{fig::firstIMG}
\end{figure*}

Originating in 2017, this task extends standard visual tracking with an additional natural language specification (also termed Vision-Language Tracking, VLT), enabling vanilla trackers to fully leverage semantic cues to improve localization performance~\cite{li2017tracking}. Specifically, Li et al. proposed an LSTM-based tracking framework with dynamic filter generation, which models text encoding and cross-modal interactions between textual and visual features, and establishes three tracking paradigms that leverage language descriptions. Wang et al. built the first TNL2K~\cite{wang2021tnl2k} benchmark dataset in 2021 and proposed a joint local-global search scheme for this task. Following this direction, DAT~\cite{wang2018describe} explicitly integrates bounding-box initialization and language descriptions for target localization, GTI~\cite{yang2020grounding} decomposes VLT into grounding, tracking, and integration, and JointNLT~\cite{zhou2023joint} further connects visual grounding and tracking within a joint framework. These efforts mainly focus on how to exploit textual information to boost tracking performance. Given the limited text generation capacity of deep models, few studies have explored dynamically adjusting textual descriptions to adapt to variations in target appearance, position, and other attributes, which hinders further performance improvement of trackers.

Driven by the booming progress of generative models and large models~\cite{wang2023MMPTMs} in particular, a wide range of computer vision tasks introduce textual inputs to address the scarcity or lack of semantic information~\cite{shi2025llmformer, jin2025LLMPAR, li2025referMOT}. Motivated by these advances, some researchers investigate adaptive updates of natural language descriptions for VLT via generative models, as shown in Fig.~\ref{fig::firstIMG} (a). More in detail, ChatTracker~\cite{sun2024chattracker} employs a multimodal large language model to refine ambiguous target descriptions with tracking feedback, DUTrack~\cite{li2025dynamic} uses BLIP~\cite{li2022blip} to generate evolving language descriptions for long-term tracking, and GLAD~\cite{luo2026glad} explores generative language-assisted fusion for more adaptive multimodal tracking.

Despite the significant improvements brought by large models, these paradigms still suffer from inherent limitations that restrict further tracking performance: 
\textit{(1) Erroneous replacement of the target object.} Most existing methods directly generate new textual descriptions through large models without explicitly distinguishing target subjects, background regions, and semantic concepts. This unstructured updating mechanism easily mismodifies or replaces the semantic information of the real target, resulting in target drift during tracking. 
\textit{(2) Background-induced semantic update drift.} These models lack the ability to selectively focus on target-relevant visual cues. They are susceptible to dynamic background disturbances and scene clutter, which leads to inappropriate updates of irrelevant semantic contents. The resulting mismatches between textual descriptions and real target states ultimately cause tracking deviation. 
\textit{(3) Unreliable textual updates caused by LLM hallucinations.} Free-form text generation inevitably produces hallucinatory, redundant, or task-irrelevant semantic information. Such inaccurate updates fail to adapt to the dynamic variations of target appearance and position, and even degrade the final tracking accuracy and robustness. 
Therefore, it is natural to raise the following question: ``\textit{How to accurately parse the given language specifications and enable targeted textual updates for large models to achieve stable and consistent performance gains in visual-language tracking?}"

Different from traditional visual-language tracking methods that treat natural language descriptions as fixed global embeddings and adopt coarse global text generation for online updating, this work argues that stable and reliable language guidance should be derived from fine-grained, structured, and target-aware semantic evolution. More in detail, this paper proposes a novel progressive structured semantic updating paradigm for vision-language tracking. Instead of directly updating the entire sentence, our method first decomposes the raw language query into explicit semantic components, including target identity, appearance concepts, and background context, via language dependency parsing. Based on the pre-trained Qwen-VL model~\cite{wang2024qwen2VL}, each semantic component is refined to enhance semantic accuracy and eliminate ambiguous or misleading descriptions. Furthermore, we design a target-conditioned Top-K visual modulation strategy to adaptively update appearance concepts according to real-time visual observations from the search region, while strictly preserving the invariance of target identity to avoid tracking drift. By integrating the structured and dynamically optimized language representation into the tracking backbone, our method achieves more accurate, robust, and consistent cross-modal matching, effectively overcoming the semantic-visual misalignment and hallucination problems existing in previous LLM-based tracking schemes. An overview of our proposed new VLT framework can be found in Fig.~\ref{fig::firstIMG} (b) and Fig.~\ref{framework}.

To sum up, the core contributions of this paper can be summarized as follows: 

$\bullet$ We propose a novel visual-language tracking framework guided by targeted natural language specification parsing and updating, which fully leverages the text generation capability of large multi-modal models, and greatly alleviates the mismatch between continuously changing target appearance during tracking and static text descriptions.

$\bullet$ We design a dependency parsing module that decomposes the initial text description into target, concept, and background components, enabling more accurate text-guided assistance and dynamic updates in the tracking process. 

$\bullet$ Extensive experiments on multiple vision-language tracking benchmark datasets (i.e., TNL2K~\cite{wang2021tnl2k}, TNLLT~\cite{wang2025reasoningtrack}, LaSOT~\cite{fan2019lasot}, OTB-LANG~\cite{li2017tracking}) fully validated the effectiveness of our proposed modules for the VLT task.

\section{Related Works} 

In this section, we will review the related works from Vision-Language Tracking and Large Foundation Models. More related works can be found from the following surveys~\cite{wang2023MMPTMs} and paper list~\footnote{\url{https://github.com/wangxiao5791509/Single_Object_Tracking_Paper_List}}. 

\subsection{Vision-Language Tracking}  

Vision-language tracking (VLT) aims to localize a target object throughout a video using both the initial bounding box and a natural language description. Compared with conventional visual tracking, language provides high-level semantic cues that are particularly useful under appearance variation, occlusion, and distractors with similar visual patterns. Early studies introduced language as an auxiliary modality for tracking. The origin of this research direction can be traced back to the pioneering work of Li et al.~\cite{li2017tracking}, which first introduced sentence-level descriptions into the object tracking task and demonstrated the potential of natural language as an auxiliary cue for target localization. Following this line of research, Wang et al.~\cite{wang2018describe} proposed DAT, which explicitly utilizes both the initial target bounding box and the corresponding language description to guide target localization, further validating that textual semantics can serve as an effective complement to visual templates. GTI~\cite{yang2020grounding} decomposed the task into grounding, tracking, and integration. To promote systematic evaluation in this area, Wang et al.~\cite{wang2021towards} established TNL2K, a large-scale benchmark for vision-language tracking, which provides standardized data and evaluation protocols and has substantially advanced subsequent research. With the development of multimodal learning, recent VLT methods have focused on stronger cross-modal interaction and unified modeling. Guo et al.~\cite{guo2022divert} proposed a unified adaptive visual-language representation for robust tracking without relying on heavy Transformer designs. JointNLT~\cite{zhou2023joint} connected visual grounding and tracking within a joint framework, while All-in-One~\cite{zhang2023all} and MMTrack~\cite{zheng2023toward} further advanced unified multimodal modeling for vision-language tracking. 

More recent studies have shifted their focus from simply introducing language cues to designing more effective multimodal interaction mechanisms. In particular, the context-aware integration framework jointly models temporal visual templates and linguistic expressions for end-to-end target decoding~\cite{shao2024context}, while UVLTrack~\cite{ma2024unifying} develops a modality-unified feature extractor to bridge visual tracking and vision-language tracking within a shared representation space. Recent studies have also begun to address the limitations of static language guidance in long-term tracking. ChatTracker~\cite{sun2024chattracker} leverages a multimodal large language model to refine ambiguous target descriptions, improving semantic guidance for tracking. DUTrack~\cite{li2025dynamic} introduces dynamic multi-modal reference updates and uses BLIP~\cite{li2022blip} to generate updated language descriptions as the target state evolves. Very recent methods further investigate dynamic multimodal adaptation from different perspectives, such as time-evolving multimodal state modeling in MambaVLT~\cite{liu2025mambavlt}, target-context alignment in ATCTrack~\cite{feng2025atctrack}, and generative language-assisted fusion in GLAD~\cite{luo2026glad}. Despite these advances, most existing methods still rely on coarse or implicit language updates and do not explicitly distinguish target-relevant semantics from redundant or background-related textual cues. In contrast, our work focuses on tracking-aware refinement of linguistic descriptions, aiming to preserve useful target semantics while suppressing misleading language during tracking.

\subsection{Large Foundation Model}

Large foundation models have recently achieved remarkable progress in visual perception, language understanding, and cross-modal reasoning. Before the recent success of MLLMs, multimedia studies had already explored cross-modal representation learning and fine-grained semantic alignment, including bidirectional cross-modal retrieval~\cite{he2016crossretrieval}, natural-language-query-based person search~\cite{zha2020personSearch}, text-video local alignment~\cite{wang2022align}, and attribute-oriented semantic modeling~\cite{abdulnabi2015multitask,Fan2022CGCN}. These studies show that fine-grained semantic matching is important for bridging visual observations and textual descriptions. More recently, foundation models such as CLIP~\cite{radford2021learning}, DINOv2~\cite{oquab2023dinov2}, SAM~\cite{kirillov2023segment}, and SAM 2~\cite{ravi2025sam} have provided transferable representations for open-vocabulary recognition, dense prediction, and promptable segmentation. In the vision-language domain, BLIP~\cite{li2022blip}, Florence-2~\cite{xiao2024florence}, LLaVA-OneVision~\cite{li2024llava}, and Qwen2.5-VL~\cite{bai2025qwen2} further demonstrate strong image-text understanding, visual grounding, instruction-following, and reasoning abilities. These advances indicate that foundation models can provide rich semantic priors for downstream vision-language tasks, including target tracking.

Motivated by these capabilities, recent studies have begun to explore foundation models and MLLMs for visual tracking. ChatTracker~\cite{sun2024chattracker} uses MLLMs to refine ambiguous target descriptions with tracking feedback, while DUTrack~\cite{li2025dynamic} employs BLIP to generate dynamic language descriptions for long-term tracking. R1-Track~\cite{wang2025r1} adapts Qwen2.5-VL to visual object tracking through supervised fine-tuning and reinforcement learning, and other works investigate MLLM-based global localization or trajectory understanding~\cite{wang2025vptracker,liao2026llmtrack}. Although these methods demonstrate the promise of foundation models for tracking, directly relying on MLLMs for localization or free-form language generation may introduce considerable computation and unstable semantic updates. Different from these methods, our work leverages foundation-model knowledge within a structured triplet space, aiming to obtain more reliable tracking-aware language refinement for vision-language tracking.

\section{Our Proposed Approach}

\subsection{Overview}

\begin{figure*}
\center
\includegraphics[width=6.5in]{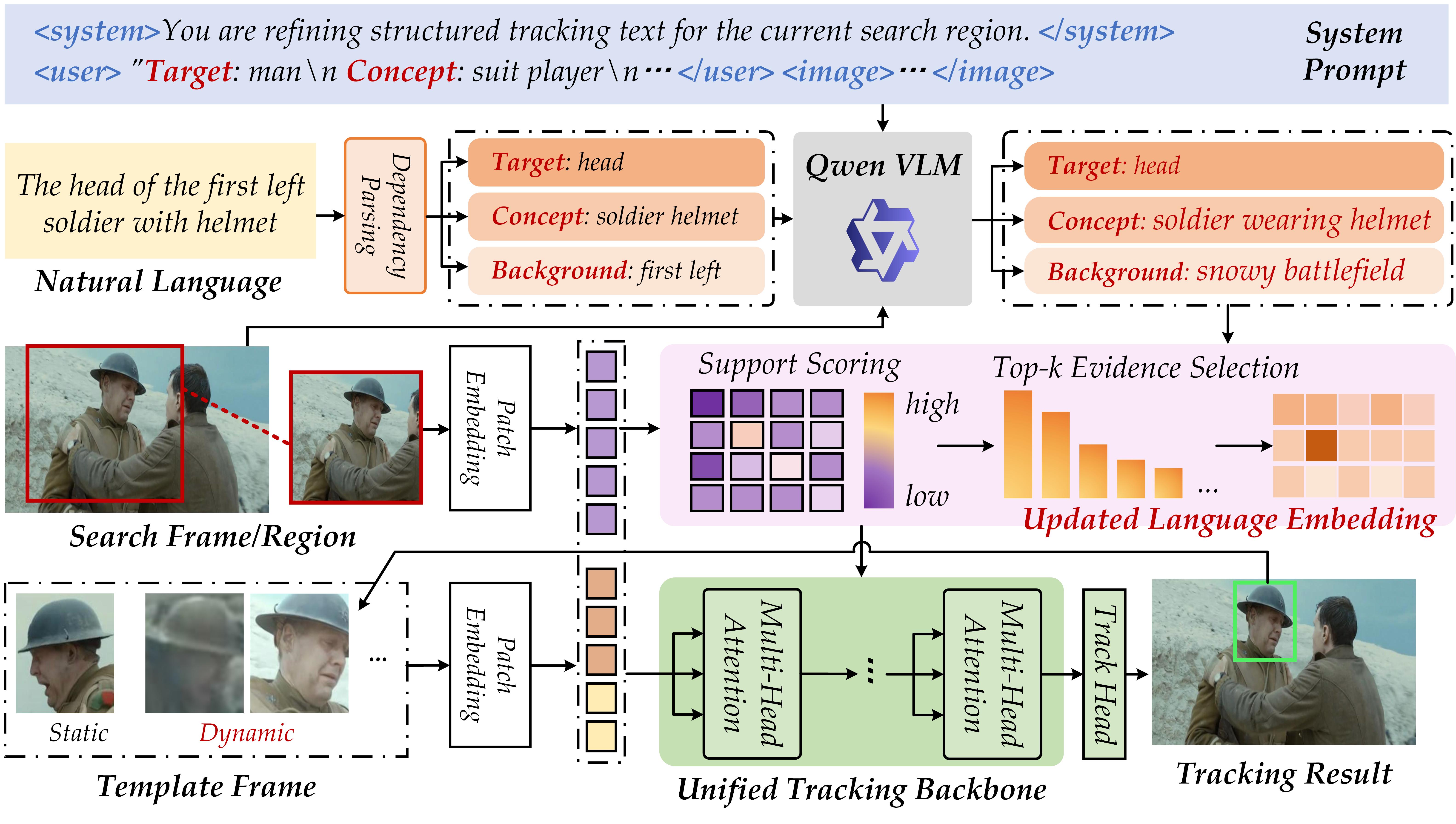} 
\caption{\textbf{An overview of the proposed fine-grained text update guided tracking framework.} Given template and search frames with a language query, we first use a dependency-parsing (DP) module to decompose the query into a structured triplet (target, concept, background). The triplet is refined by Qwen to improve semantic quality. Then, target-conditioned Top-K visual evidence is used to update concept spans via span-local cross-attention writeback, while the target identity and background fields remain unchanged. The resulting structured language representation is fused with visual features in a unified tracking backbone to produce final tracking predictions.}
\label{framework}
\end{figure*}

Given a template frame, a search frame, and a natural language query, vision-language tracking aims to localize the target object in the search frame according to both visual evidence and language guidance. Unlike conventional formulations that directly encode the whole sentence as a single text representation, we focus on a more fine-grained text update setting, where the language description is explicitly decomposed and progressively refined before being fed into the tracker. As illustrated in Figure~\ref{framework}, our framework follows a three-stage progressive pipeline. Given a raw language query, a dependency parsing module first decomposes it into a structured triplet covering target identity, appearance concepts, and background context. A Qwen-based refinement module then improves the semantic quality of each field. Finally, a target-conditioned Top-K visual modulation stage updates the concept field according to the current search-region evidence, while keeping the target identity stable. The resulting structured representation is fused with visual features in a unified tracking backbone for target localization. 

\subsection{Fine-grained Text Update Guided Tracking Framework}

\textbf{Fine-grained Text Update Paradigm.}
Following the overall architecture in Figure~\ref{framework}, we formulate our method as a progressive language-state update framework for vision-language tracking. Instead of encoding the input query only once and directly using it throughout the whole tracking process, we treat the structured triplet as an intermediate textual state that can be successively refined and adapted to the current visual observation. Under this formulation, language guidance is no longer static, but evolves together with the tracking process in a fine-grained manner.

Specifically, let $\mathcal{T}=\{t^{\mathrm{tar}}, t^{\mathrm{con}}, t^{\mathrm{bg}}\}$ denote the parsed triplet obtained from the raw query, where the three fields represent target identity $t^{\mathrm{tar}}$, target-related concepts $t^{\mathrm{con}}$, and background context $t^{\mathrm{bg}}$, respectively. Based on this structured state, we first apply a language-side refinement module to obtain
\begin{equation}
\hat{\mathcal{T}} = \Phi_{\mathrm{qwen}}(\mathcal{T}),
\end{equation}
where $\Phi_{\mathrm{qwen}}(\cdot)$ denotes the Qwen-based triplet refinement operator. In practice, the parsed triplet is serialized into a structured instruction, where the target, concept, and background fields are explicitly marked and fed into the Qwen refiner. The refiner is required to preserve the target identity, remove redundant or misleading expressions, and rewrite the concept and background fields into more compact tracking-oriented descriptions. This stage mainly improves the semantic quality of the triplet representation before visual-conditioned concept updating.

After that, we further introduce a visual-conditioned update stage to align the refined triplet with the current search frame. Formally, given the search-region visual tokens $\mathbf{z}_s$, the refined triplet is updated as
\begin{equation}
\tilde{\mathcal{T}} = \Phi_{\mathrm{topk}}(\hat{\mathcal{T}}, \mathbf{z}_s),
\end{equation}
where $\Phi_{\mathrm{topk}}(\cdot)$ denotes the target-conditioned sparse visual modulation operator. Different from conventional text-visual fusion that treats all textual components uniformly, our design explicitly preserves the target field as a stable identity anchor and only performs fine-grained visual updating on the concept field. This is because the concept field mainly captures appearance-sensitive attributes, which should be dynamically adapted according to the current visual evidence, whereas the target identity should remain semantically stable.

More concretely, the update process can be written as
\begin{equation}
\tilde{\mathcal{T}}=
\left\{
\hat{t}^{\mathrm{tar}},
\;
\Psi(\hat{t}^{\mathrm{con}}, \mathbf{z}_s ; \hat{t}^{\mathrm{tar}}),
\;
\hat{t}^{\mathrm{bg}}
\right\},
\end{equation}
where $\Psi(\cdot)$ denotes the concept refinement operator conditioned on the target field and the search-region visual evidence. Specifically, the target field is used as an identity anchor to measure the relevance between the target semantics and search-region visual tokens, from which a sparse set of Top-K target-related evidence is selected. The selected visual evidence is then written back only to the concept span, enabling the concept description to be adjusted according to the most relevant visual cues in the current frame, while the target field remains unchanged as the semantic anchor of the whole triplet.

Accordingly, our framework formulates language guidance as a progressive state transition process from parsed triplets to semantically refined triplets and finally to visually updated triplets. This design preserves stable target identity while adaptively refining appearance-sensitive concept descriptions, leading to more precise guidance for vision-language tracking.

\textbf{Tracking Inference Workflow.}
At each inference step, the tracker takes as input a template frame $I_t$, a search frame $I_s$, and a language query $Q$. Following the fine-grained text update paradigm described above, the query is first converted into a structured triplet and then refined by the Qwen module, yielding the refined textual state $\hat{\mathcal{T}}$. Based on $\hat{\mathcal{T}}$, the tracker further performs target-conditioned visual modulation on the current search frame to obtain the updated triplet $\tilde{\mathcal{T}}$, which is subsequently used for multimodal fusion and target state prediction.

Let $\mathbf{z}_t \in \mathbb{R}^{N_t \times d}$ and $\mathbf{z}_s \in \mathbb{R}^{N_s \times d}$ denote the template and search visual tokens extracted by the tracking backbone, respectively, where $N_t$ and $N_s$ are the numbers of template and search tokens, and $d$ is the feature dimension. Let $\mathbf{e}^{\mathrm{tar}} \in \mathbb{R}^{d}$ be the mean-pooled text embedding of the target field $\hat{t}^{\mathrm{tar}}$. We use it as an identity anchor to score each search-region token via cosine similarity:
\begin{equation}
a_i = \frac{\mathbf{e}^{\mathrm{tar}} \cdot \mathbf{z}_{s,i}}{\|\mathbf{e}^{\mathrm{tar}}\|\,\|\mathbf{z}_{s,i}\|}, \quad i=1,\ldots,N_s,
\end{equation}
where $\mathbf{a} \in \mathbb{R}^{N_s}$ collects all scores. Based on $\mathbf{a}$, we select the Top-K most informative search tokens as sparse visual evidence,
\begin{equation}
\mathbf{z}_s^{K} = \mathrm{TopK}(\mathbf{z}_s, \mathbf{a}, K).
\end{equation}
The selected evidence is then written back to the concept field through a span-local cross-attention operator. Let $\mathbf{E}^{\mathrm{con}} \in \mathbb{R}^{L \times d}$ denote the token embeddings of the concept span $\hat{t}^{\mathrm{con}}$ with length $L$. The updated concept embedding is computed as
\begin{equation}
\tilde{\mathbf{E}}^{\mathrm{con}} = \mathbf{E}^{\mathrm{con}} + \mathrm{CrossAttn}(\mathbf{E}^{\mathrm{con}},\, \mathbf{z}_s^{K},\, \mathbf{z}_s^{K}),
\end{equation}
where $\mathrm{CrossAttn}(Q,K,V)$ denotes standard scaled dot-product attention with $\mathbf{E}^{\mathrm{con}}$ as query and $\mathbf{z}_s^{K}$ as key and value. The resulting $\tilde{\mathbf{E}}^{\mathrm{con}}$ is decoded back to the concept field $\tilde{t}^{\mathrm{con}}$, while the target and background fields remain unchanged. Accordingly, the final updated triplet is expressed as
\begin{equation}
\tilde{\mathcal{T}} =
\left\{
\hat{t}^{\mathrm{tar}},
\tilde{t}^{\mathrm{con}},
\hat{t}^{\mathrm{bg}}
\right\}.
\end{equation}

After the fine-grained text update, the tracker fuses the updated triplet $\tilde{\mathcal{T}}$ with the template and search visual features for target localization. The final prediction is given by
\begin{equation}
\mathbf{y} = \mathcal{H}(\mathbf{z}_t, \mathbf{z}_s, \tilde{\mathcal{T}}),
\end{equation}
where $\mathcal{H}(\cdot)$ denotes the unified tracking network together with the prediction head, and $ \mathbf{y}$ is the predicted target state in the current frame. In this way, the tracker dynamically updates appearance-sensitive concept descriptions according to the current visual observation, while preserving the semantic stability of the target identity throughout the inference process.

\subsection{DP Triplet Parsing Sub-Network}


Given a raw language query $Q=\{w_1,\ldots,w_n\}$, the goal of the DP triplet parsing sub-network is to convert the original sentence into a structured triplet representation
\begin{equation}
\mathcal{T}=\{t^{\mathrm{tar}}, t^{\mathrm{con}}, t^{\mathrm{bg}}\},
\end{equation}
where $t^{\mathrm{tar}}$, $t^{\mathrm{con}}$, and $t^{\mathrm{bg}}$ denote the target, concept, and background fields, respectively. Instead of directly relying on a large language model to produce such decomposition, we first introduce a lightweight dependency-based parser to provide an explicit structural initialization for subsequent refinement.

Concretely, we adopt a biaffine dependency parser~\cite{dozat2017deep} trained on the Universal Dependencies English Web Treebank (UD-EWT)~\cite{nivre2016universal,silveira2014gold}. For an input sentence, each token is represented by the sum of a word embedding and a universal POS embedding. Let $\mathbf{E}^{w}\in\mathbb{R}^{n\times d}$ and $\mathbf{E}^{p}\in\mathbb{R}^{n\times d}$ denote the corresponding word and POS embeddings. The parser input is defined as
\begin{equation}
\mathbf{X}=\mathbf{E}^{w}+\mathbf{E}^{p}.
\end{equation}

The token sequence is then encoded by a multi-layer Transformer encoder with positional encoding,
\begin{equation}
\mathbf{H}=\mathrm{Encoder}(\mathbf{X}),
\end{equation}
where $\mathbf{H}\in\mathbb{R}^{n\times d}$ denotes the contextualized token features.

Based on $\mathbf{H}$, we construct separate head and modifier representations for dependency arc prediction and dependency label prediction. Specifically, two feed-forward projections are applied before biaffine scoring:
\begin{equation}
\mathbf{H}^{\mathrm{arc}}_{h},\mathbf{H}^{\mathrm{arc}}_{m}
= f_{\mathrm{arc}}(\mathbf{H}), \quad
\mathbf{H}^{\mathrm{rel}}_{h},\mathbf{H}^{\mathrm{rel}}_{m}
= f_{\mathrm{rel}}(\mathbf{H}),
\end{equation}
where $f_{\mathrm{arc}}(\cdot)$ and $f_{\mathrm{rel}}(\cdot)$ denote task-specific projection layers. The dependency arc score between token $i$ and token $j$ is computed by a biaffine form
\begin{equation}
s^{\mathrm{arc}}_{i,j}
=
(\mathbf{h}^{\mathrm{arc}}_{m,i})^{\top}\mathbf{W}_{\mathrm{arc}}\mathbf{h}^{\mathrm{arc}}_{h,j}
+
b^{\mathrm{pos}}_{i,j},
\end{equation}
where $\mathbf{W}_{\mathrm{arc}}$ is the arc scoring parameter and $b^{\mathrm{pos}}_{i,j}$ is a relative positional bias. The dependency label score is computed in a similar way through a label biaffine classifier,
\begin{equation}
\mathbf{s}^{\mathrm{rel}}_{i,j}
=
\mathrm{Biaffine}_{\mathrm{rel}}
(\mathbf{h}^{\mathrm{rel}}_{m,i}, \mathbf{h}^{\mathrm{rel}}_{h,j}).
\end{equation}

The parser is trained with standard arc prediction and dependency relation classification objectives.

After obtaining the predicted dependency tree, we convert it into a triplet representation through a lightweight rule-based extraction procedure. We first select the target anchor from the nominal tokens according to dependency priority. In particular, noun or proper-noun tokens attached to the sentence root are preferred, followed by nominal subjects and then direct or indirect objects. Once the anchor token is determined, we expand it with its local nominal modifiers, including adjectival modifiers, compounds, numeric modifiers, possessives, and flat structures, to form the target phrase $t^{\mathrm{tar}}$.

The concept field $t^{\mathrm{con}}$ is then constructed from the remaining nominal tokens that do not belong to the target phrase. During this step, locator-type words and ordinal indicators are excluded, since they are more closely related to spatial context than target appearance. Finally, the background field $t^{\mathrm{bg}}$ is extracted from location-sensitive expressions, such as directional words, ordinal phrases, and spatial patterns like ``from left to right''. Duplicate entries are removed while preserving their original order.

The parsed output is represented as a structured triplet 
$
\mathcal{T} =
\{t^{\mathrm{tar}}, t^{\mathrm{con}}, t^{\mathrm{bg}}\},
$
where $t^{\mathrm{tar}}$ denotes the target anchor, and $t^{\mathrm{con}}$ and $t^{\mathrm{bg}}$ denote the concept and background fields, respectively. This triplet is used as the initial structured language state for the subsequent Qwen refinement and target-conditioned concept update modules.

\begin{figure}
\centering
\includegraphics[width=\linewidth]{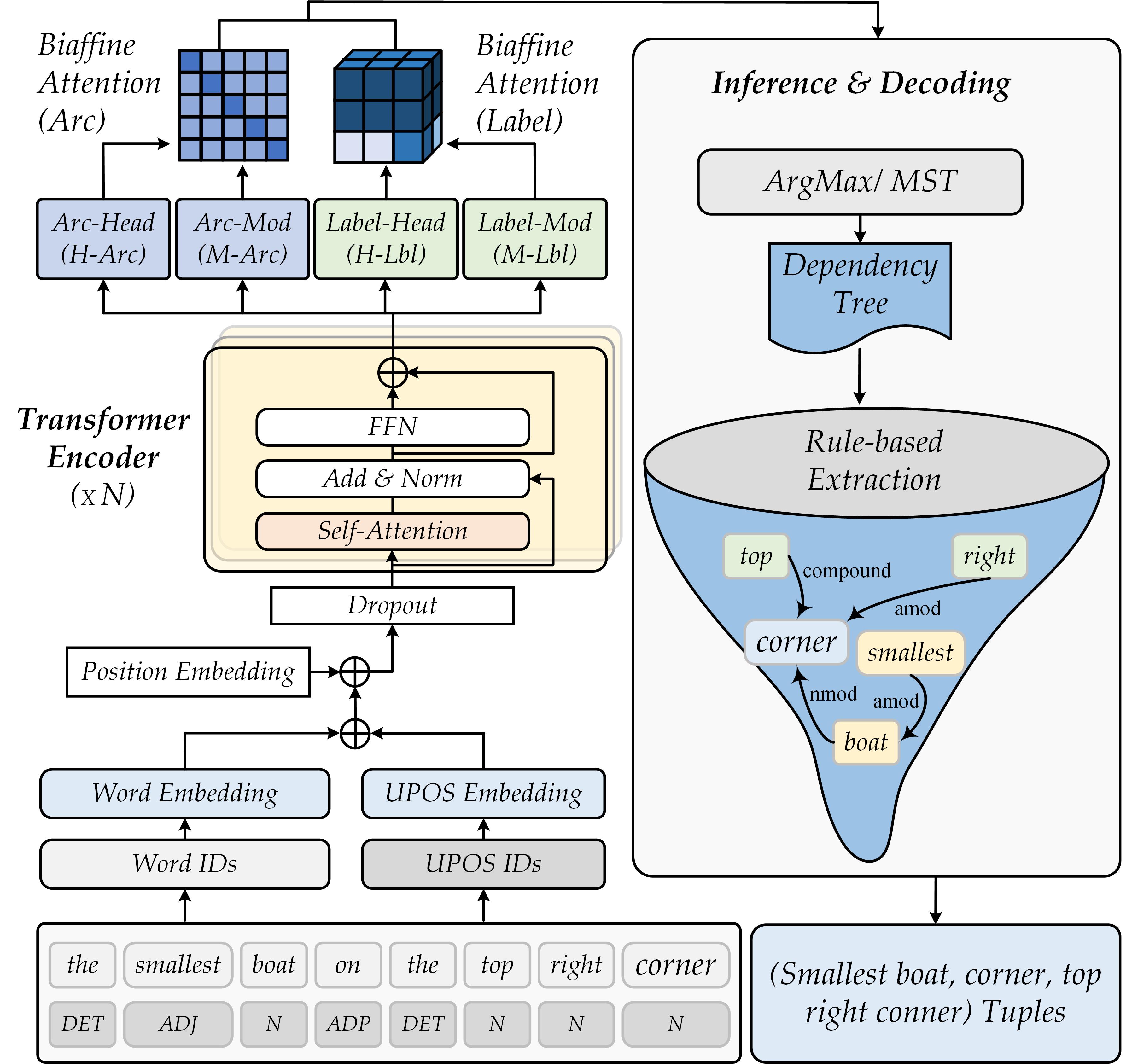}
\caption{\textbf{Architecture of the dependency parsing module.} The module encodes the raw query with word and POS embeddings, predicts dependency arcs and relation labels through biaffine classifiers, and converts the resulting dependency tree into a structured triplet consisting of target, concept, and background fields.}  
\label{fig:dp_module}
\end{figure}

\subsection{Qwen Refinement Sub-Network}

\textbf{Triplet Refinement Strategy.~} 
Given the parsed triplet $\mathcal{T}=\{t^{\mathrm{tar}}, t^{\mathrm{con}},
t^{\mathrm{bg}}\}$ produced by the DP triplet parsing sub-network, we further introduce a Qwen-based refinement module to improve the semantic quality of the structured fields before visual grounding. The motivation is that dependency-based decomposition provides an explicit structural initialization, but the resulting fields may still contain noisy expressions, incomplete attributes, or linguistically unnatural fragments that are suboptimal for downstream tracking.

To address this issue, we formulate triplet refinement as a structured text-to-text transformation process. Specifically, the Qwen module takes the parsed triplet as input and outputs a refined triplet
$\hat{\mathcal{T}}=\{\hat{t}^{\mathrm{tar}}, \hat{t}^{\mathrm{con}}, \hat{t}
^{\mathrm{bg}}\}$,
where the target, concept, and background fields are refined in a role-aware manner. In our design, the refinement mainly focuses on improving the concept and background descriptions, while preserving the target identity as much as possible to avoid semantic drift. As a result, the refined triplet is more compact, discriminative, and better aligned with the language requirements of the tracking model.

Formally, the refinement process is written as
\begin{equation}
\hat{\mathcal{T}}=\Phi_{\mathrm{qwen}}(\mathcal{T}),
\end{equation}
where $\Phi_{\mathrm{qwen}}(\cdot)$ denotes the Qwen refinement operator. Different from directly generating a new raw sentence, our module operates on the structured triplet space, which preserves role-specific semantics and provides a more controllable interface for subsequent target-conditioned visual updating.

\textbf{Tracking-aware Supervised Fine-tuning.}
To make the refinement module more suitable for vision-language tracking, we perform tracking-aware supervised fine-tuning on Qwen. Instead of treating triplet refinement as a purely linguistic rewriting task, we construct supervision by explicitly considering its effect on downstream tracking performance.

Concretely, for each parsed triplet $\mathcal{T}$, we generate $M$ refined candidates $\{\hat{\mathcal{T}}^{(m)}\}_{m=1}^{M}$ via beam search with beam size $M$. Each candidate is evaluated by running the tracker on a held-out set of frames and computing the average overlap (AUC) between the predicted boxes and ground-truth annotations. Candidates whose AUC exceeds a threshold $\tau$ are retained as positive supervision targets, while the rest are discarded. The Qwen refiner is then fine-tuned with a standard next-token cross-entropy loss on the positive targets:
\begin{equation}
\mathcal{L}_{\mathrm{sft}} = -\sum_{m \in \mathcal{P}} \sum_{k} \log p_{\theta}(\hat{w}^{(m)}_k \mid \hat{w}^{(m)}_{<k},\, \mathcal{T}),
\end{equation}
where $\mathcal{P}$ denotes the set of positive candidate indices and $\hat{w}^{(m)}_k$ is the $k$-th output token of the $m$-th candidate. In practice, we adopt LoRA~\cite{hu2022lora} for parameter-efficient fine-tuning to keep the training cost manageable. This design enables the refinement module to learn tracking-oriented rewriting behavior without changing the overall triplet-based language interface.

\textbf{Integration with the Tracking Pipeline.}
During inference, the Qwen refinement sub-network is placed immediately after the DP triplet parsing stage and before the target-conditioned visual update stage. Given an input query $Q$, the tracker first obtains the parsed triplet $\mathcal{T}$, and then feeds it into the Qwen refiner to produce the refined triplet $\hat{\mathcal{T}}$. The refined triplet is subsequently passed to the target-conditioned concept update module, where the concept field is further adjusted according to the current search-region visual evidence.

Under this design, the Qwen refinement stage and the visual refinement stage play complementary roles in the overall pipeline. The former mainly improves the semantic quality and structural consistency of the triplet representation, while the latter further adapts the appearance-sensitive concept field to the current frame. As a result, the tracker receives a structured language representation that is both semantically refined and visually grounded, which provides more precise guidance for downstream multimodal interaction and target localization.

Importantly, the Qwen refinement sub-network operates entirely within the triplet space and does not alter the downstream tracking architecture. It serves as an intermediate structured language enhancement module between dependency-based parsing and target-conditioned concept updating, making the overall fine-grained text update pipeline more stable and effective.

\subsection{Head and Loss Function}

\textbf{Tracking Head.~}
After the fine-grained text update process, the updated triplet $\tilde{\mathcal{T}}=\{\hat{t}^{\mathrm{tar}}, \tilde{t}^{\mathrm{con}}, \hat{t}^{\mathrm{bg}}\}$ is encoded into three separate text embeddings. The target and concept embeddings provide foreground guidance, while the background embedding $\hat{t}^{\mathrm{bg}}$ supplies spatial context (e.g., scene layout and relative position) that helps suppress distractors in cluttered environments. All three embeddings are concatenated and fused with the template and search visual features in the unified tracking backbone via cross-attention. Based on the resulting multimodal representation, we adopt the same tracking head as the baseline tracker to predict the target state in the current search frame. Following the standard center-based formulation, the head produces a target score map together with the corresponding box regression outputs for target localization.

\textbf{Training Objectives.~} 
The training objective is defined on the final tracking prediction. The regression loss $\mathcal{L}_{\mathrm{reg}}$ combines an $\ell_1$ term and a GIoU term~\cite{rezatofighi2019generalized} on the predicted bounding box, and the localization loss~\cite{lin2017focal} $\mathcal{L}_{\mathrm{loc}}$ is a focal loss applied to the predicted score map. Formally,
\begin{equation}
\mathcal{L}_{\mathrm{total}}
=
\lambda_{1}\mathcal{L}_{\mathrm{loc}}
+
\lambda_{2}\mathcal{L}_{\ell_1}
+
\lambda_{3}\mathcal{L}_{\mathrm{GIoU}},
\end{equation}
where $\lambda_1$, $\lambda_2$, and $\lambda_3$ are balancing weights. The proposed fine-grained text update modules are trained end-to-end through this objective to enhance multimodal target localization.

\section{Experiments} 

\subsection{Datasets and Evaluation Metric} 
We evaluate the proposed method on four language-guided tracking benchmarks, including TNLLT~\cite{wang2025reasoningtrack}, TNL2K~\cite{wang2021tnl2k}, LaSOT~\cite{fan2019lasot}, and OTB99-Lang~\cite{li2017tracking}. 
Following the standard one-pass evaluation protocol of each benchmark, we report precision-based and overlap-based metrics. Specifically, TNLLT is evaluated with Precision (PR), Normalized Precision (NPR), and Success Rate (SR); TNL2K is evaluated with Area Under the Curve (AUC) and PR; LaSOT is evaluated with AUC, NPR, and PR; and OTB99-Lang is evaluated with PR and AUC. All reported scores are higher-the-better. 


\begin{figure*}[!htp]
\centering
\includegraphics[width=\linewidth]{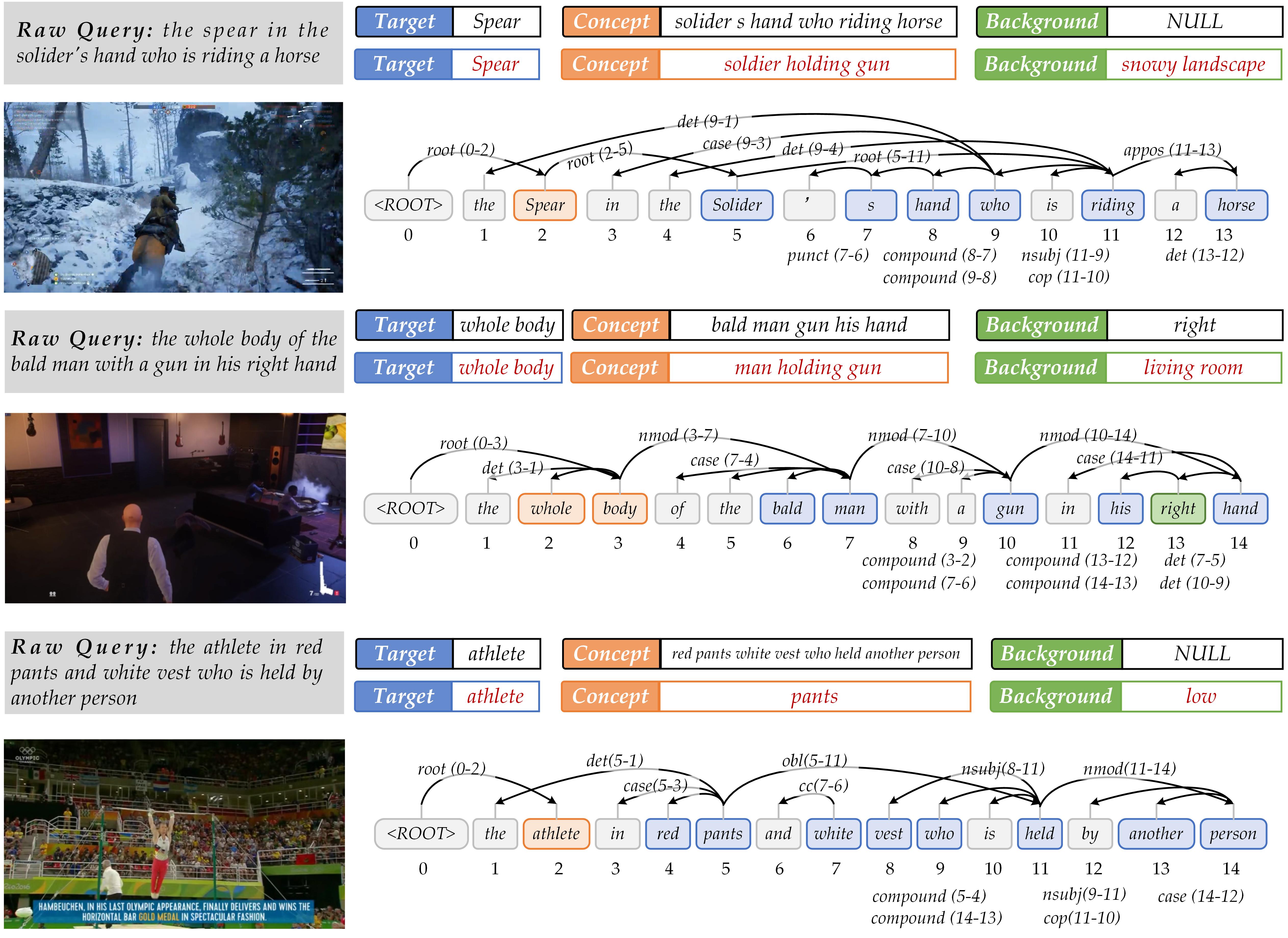}
\caption{\textbf{Visualization of dependency-based triplet parsing.} For each example, the raw query is parsed into token-level dependency relations and converted into a structured triplet consisting of target, concept, and background fields. The upper triplet in each row shows the initial parsing result, while the lower triplet shows the refined language state used by our tracker. The visualization illustrates that the proposed parsing process preserves stable target identity while reorganizing appearance, interaction, and contextual cues into role-specific fields.}
\label{LDP}
\end{figure*}

\subsection{Implementation Details} 

\textbf{Tracker Architecture.~}
Our framework is built on a unified vision-language tracking architecture with an ITPN-Base visual backbone and a center-based prediction head. The template and search resolutions are set to $128\times128$ and $256\times256$, respectively, with template and search region factors of 2.0 and 4.0. During training, we use three template frames and two search regions for each sample, while four template frames are used at test time. The backbone uses a stride of 16 and adopts the same concatenation-style multimodal interaction setting throughout all experiments.

\textbf{Triplet Parsing and Qwen Refinement.~}
For the language branch, each raw query is first decomposed by the DP module into three structured fields, namely target, concept, and background. We use the parsed-text setting together with token-level concatenation, and the maximum total text length is set to 32. The structured fields are fed into the tracker in a role-aware manner rather than being merged into a single unstructured sentence. For fine-grained text updating, we only apply visual-conditioned writeback to the concept field, while keeping the target identity unchanged. The target-conditioned evidence selection module uses cosine-based support scoring with temperature $\tau=0.07$, and the final model adopts a sparse Top-K setting with $K=8$.

For the Qwen refinement module, we initialize the refiner from Qwen2.5-VL-3B-Instruct~\cite{bai2025qwen2} and perform tracking-aware LoRA fine-tuning on the triplet refinement task. The Qwen module takes the parsed triplet and search area as input and outputs a semantically refined structured triplet before the target-conditioned concept update stage. Unless otherwise specified, the refined triplet is further processed by the same Top-K concept writeback module before multimodal fusion.

\textbf{Training and Inference Settings.~}
We optimize the tracker with AdamW~\cite{loshchilov2019decoupled} using a batch size of 8, an initial learning rate of $1\times10^{-4}$, and a weight decay of $1\times10^{-4}$. The learning rate is decayed at epoch 20, and the gradient clipping norm is set to 0.1. For box regression, the weights of the $\ell_1$ loss and GIoU loss are set to 5.0 and 2.0, respectively. We train the model for 30 epochs on TNLLT, TNL2K, LaSOT, and OTB99-Lang. The number of sampled training pairs per epoch is set to 30,000 on TNLLT, TNL2K, and LaSOT, and 12,000 on OTB99-Lang, with the corresponding validation samples set to 6,000 and 2,400. During inference, the tracker first performs structured text refinement and then applies target-conditioned sparse concept updating before producing the final target prediction.

\subsection{Comparison on Public Benchmark Datasets} 

$\bullet$ \textbf{Results on TNLLT Dataset.~} 
Table~\ref{benchmark} reports the comparison results on TNLLT~\cite{wang2025reasoningtrack}. TNLLT contains 200 long-term sequences with natural-language queries and absent-target annotations, covering challenging cases such as target disappearance, reappearance, severe appearance changes, and distractor interference. Therefore, it provides a suitable benchmark for evaluating robust language-guided tracking. Our method achieves 75.0 PR, 78.2 NPR, and 64.5 SR, obtaining the best PR and SR among the compared trackers while maintaining competitive NPR. Compared with DUTrack, our method improves PR, NPR, and SR by 2.5, 2.4, and 1.7 points, respectively. Compared with ReasoningTrack, our method further improves the three metrics by 0.9, 1.2, and 0.6 points. Although SDTrack reports a higher NPR, its much lower PR and SR indicate that its advantage mainly comes from normalized center accuracy rather than precise absolute localization or bounding-box overlap. These results indicate that structured triplet refinement and target-conditioned concept updating provide more adaptive language guidance for long-term tracking.

\begin{figure*}[!htp]
\centering
\includegraphics[width=\linewidth]{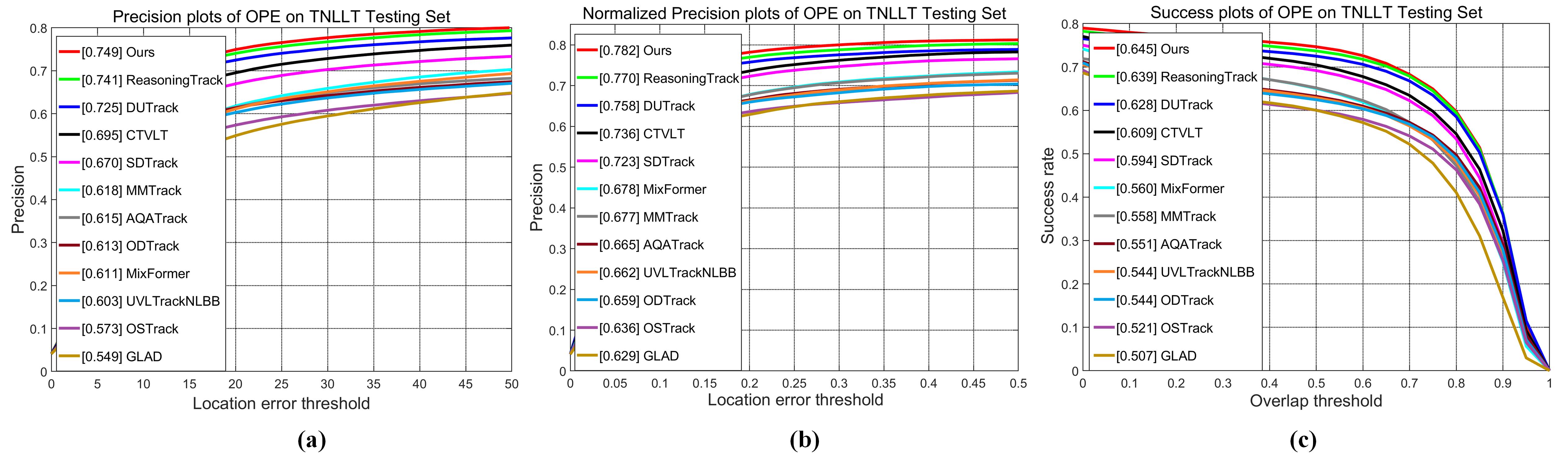}
\caption{\textbf{Tracking results on TNLLT.} PR, NPR, and SR curves are reported to compare different trackers under varying evaluation thresholds. Our method achieves consistently strong performance, showing the benefit of fine-grained text updating for robust vision-language tracking.}
\label{SRPRNPR}
\end{figure*}

Figure~\ref{SRPRNPR} further visualizes the PR, NPR, and SR curves on TNLLT. The curves show that our method maintains strong performance across different evaluation thresholds, further supporting the effectiveness of the proposed fine-grained text update strategy.

\begin{table}
\center
\scriptsize   
\setlength{\tabcolsep}{3.8pt}
\renewcommand{\arraystretch}{1.1}
\caption{Overall Tracking Performance on TNLLT Dataset. } 
\label{benchmark}
\resizebox{\columnwidth}{!}{ 
\begin{tabular}{l|l|l|ccc}
\hline \toprule [0.5 pt]
\textbf{Trackers} & \textbf{Source} & \textbf{Type} & \textbf{PR}  &\textbf{NPR}   &\textbf{SR}  \\
\hline
\textbf{OSTrack~\cite{ye2022ostrack}} & ECCV 2022 & BB & 57.3 & 63.6 & 52.1  \\
\textbf{MixFormer~\cite{cui2022mixformer}} & CVPR 2022 & BB & 61.1 & 67.8 & 56.0 \\
\textbf{AiATrack~\cite{gao2022aiatrack}} & ECCV 2022 & BB & 56.9 & 63.5 & 53.0 \\
\textbf{CiteTrack~\cite{citetracker}} & ICCV 2023 & BB & 55.3 & 60.8 & 50.3 \\
\textbf{ROMTrack~\cite{cai2023robust}} & ICCV 2023 & BB & 54.8 & 59.9 & 49.0 \\
\textbf{GRM~\cite{gao2023generalized}} & CVPR 2023 & BB & 55.4 & 61.3 & 50.0 \\
\textbf{ODTrack~\cite{zheng2024odtrack}} & AAAI 2024 & BB & 61.3 & 65.9 & 54.4   \\
\textbf{EVPTrack~\cite{shi2024evptrack}} & AAAI 2024 & BB & 61.3 & 65.8 & 54.4 \\
\textbf{UVLTrack~\cite{ma2024unifying}} & AAAI 2024 & BB & 58.1 & 64.1 & 53.1 \\
\textbf{AQATrack~\cite{xie2024autoregressive}} & CVPR 2024 & BB & 61.5 & 66.5 & 55.1 \\
\textbf{LMTrack~\cite{xu2025less}} & AAAI 2025 & BB &53.0 & 59.8 & 48.7 \\
\hline
\textbf{JointNLT~\cite{zhou2023joint}} & CVPR 2023 & NL & 47.9 & 55.5 & 45.0 \\
\textbf{JointNLT~\cite{zhou2023joint}} & CVPR 2023 & BL & 48.2 & 55.9 & 45.4 \\
\textbf{All-in-one~\cite{zhang2023all}} & ACM MM 2023 & BL & 54.8 & 60.6 & 49.7  \\
\textbf{MMTrack~\cite{zheng2023toward}} & TCSVT 2023 & BL & 61.8 & 67.7 & 55.8 \\
\textbf{UVLTrack~\cite{ma2024unifying}} & AAAI 2024 & NL & 57.4 & 63.1 & 51.9 \\
\textbf{UVLTrack~\cite{ma2024unifying}} & AAAI 2024 & BL & 60.3 & 66.2 & 54.4 \\
\textbf{CTVLT~\cite{feng2025enhancing}} & ICASSP 2025 & BL & 69.5 & 73.6 & 60.9  \\
\textbf{SUTrack~\cite{sutrack}} & AAAI 2025 & BL & 55.4 & 60.0 & 50.8   \\
\textbf{DUTrack~\cite{Li2025dutrack}} & CVPR 2025 & BL & 72.5 & 75.8 & 62.8 \\
\textbf{ReasoningTrack\cite{wang2025reasoningtrack}} &arXiv 2025  & BL & 74.1 & 77.0 & 63.9  \\
\textbf{GLAD}\cite{luo2026glad}   &IJCV 2026    & BL  & 51.6 & 72.6 & 51.9  \\
\textbf{SDTrack}\cite{bai2026selective}   &ESWA 2026    & BL   & 64.0  & 82.3  & 61.0   \\
\hline
\textbf{Ours} & - & BL & 75.0 & 78.2 & 64.5  \\
\hline \toprule [0.5 pt]
\end{tabular}
}
\end{table}

$\bullet$ \textbf{Results on TNL2K Dataset.~}
Table~\ref{TNL2k_results} shows the results on TNL2K~\cite{wang2021tnl2k}. TNL2K is a large-scale vision-language tracking benchmark containing 2,000 video sequences with diverse object categories, complex scenes, and rich natural-language descriptions. Its large scale and language diversity make it suitable for comprehensively evaluating cross-modal target localization ability. Our method obtains 65.1 AUC and 74.5 PR. In terms of PR, our tracker achieves the best result among the compared methods, surpassing DUTrack by 3.9 points and ReasoningTrack by 0.2 points. The AUC is slightly lower than ATSTrack and ReasoningTrack, which suggests that our fine-grained text update strategy is particularly effective for improving precise target localization, while overlap-based robustness can be further enhanced. Overall, the TNL2K results verify that our method generalizes well beyond the long-term TNLLT setting.

\begin{table}
\center
\small   
\caption{Overall tracking performance on TNL2K dataset. } 
\label{TNL2k_results}
\begin{tabular}{l|l|cccc}
\hline \toprule [0.5 pt]
\textbf{Trackers} & \textbf{Source}   & \textbf{AUC}     &\textbf{PR} \\
\hline
\textbf{OSTrack}\cite{ye2022ostrack}    &ECCV 2022           &55.9          &-        \\  
\textbf{SeqTrack}\cite{chen2023seqtrack}    &CVPR 2023          &54.9          &-      \\
\textbf{TDCL }\cite{wang2024tdcl}   &ECAI 2024      & 49.6     &  56.5 \\ 
\textbf{AQATrack}\cite{xie2024autoregressive}    &CVPR 2024           &57.8            &59.4       \\ 
\textbf{ODTrack-384}\cite{zheng2024odtrack}    &AAAI 2024          &60.9           &-         \\ 
\textbf{JointNLT}\cite{zhou2023joint}    &CVPR 2023          &56.9           &58.1        \\ 
\textbf{MMTrack-384}\cite{zheng2023toward}    &TCSVT 2023           &58.6             &59.4         \\ 
\textbf{ATTrack}\cite{ge2024consistencies}    &MM 2024          &56.9            &64.7         \\ 
\textbf{UVLTrack}\cite{ma2024unifying}    &AAAI 2024          &62.7             &65.4       \\ 
\textbf{OSDT}\cite{zhang2024one}    &TCSVT 2024           &59.3             &61.5         \\ 
\textbf{QueryNLT}\cite{shao2024context}    &CVPR 2024          &57.8             &58.7         \\ 
\textbf{DUTrack}\cite{Li2025dutrack}    &CVPR 2025           &64.9          &70.6     \\ 
\textbf{ATSTrack}\cite{zhen2025atstrack}    &arXiv 2025           &66.2           &71.3   \\
\textbf{TemTrack}\cite{xie2025robust}    &AAAI 2025           &58.8           &-   \\
\textbf{ReasoningTrack}\cite{wang2025reasoningtrack}    &arXiv 2025      & 65.3     &  74.3 \\ 
\textbf{GLAD}\cite{luo2026glad}   &IJCV 2026      & 59.7     &  62.2 \\ 
\textbf{ADTrack }\cite{zhang2026aware}   &AAAI 2026      & 65.3     &  70.8 \\ 
\textbf{SDTrack }\cite{bai2026selective}   &ESWA 2026      & 65.6     &  70.8 \\ 
\textbf{ABOTrack }\cite{hu2026vision}   &JVCIR 2026      & 58.9     &  60.1 \\ 
\textbf{SemaTrack }\cite{zhang2026sematrack}   &TVC 2026      & 65.5     &  70.2 \\ 
\textbf{RrmTrack }\cite{zhang2026rwkv}   &TMM 2026      & 62.1     &  66.0 \\ 
\hline
\textbf{Ours }   & -      & 65.1     &  74.5 \\
\hline \toprule [0.5 pt]
\end{tabular}
\end{table}

$\bullet$ \textbf{Results on LaSOT dataset.~} 
Table~\ref{LaSOT_results} reports the comparison results on LaSOT~\cite{fan2019lasot}. LaSOT is a large-scale long-term tracking benchmark containing 1,400 videos with dense frame-level annotations, long sequence duration, and diverse object categories. These properties make it a challenging testbed for evaluating tracker robustness under large appearance variations, occlusion, and distractors. Our method achieves 71.7 AUC, 83.1 NPR, and 80.3 PR, showing competitive performance among recent trackers. In particular, the strong NPR and PR results indicate that the proposed fine-grained text update framework can provide stable target localization under large-scale long-term tracking scenarios. These results suggest that our method generalizes well to challenging visual tracking benchmarks with complex appearance variations.

\begin{table}
\center
\small   
\caption{Overall tracking performance on LaSOT dataset. } 
\label{LaSOT_results}
\begin{tabular}{l|l|ccccc}
\hline \toprule [0.5 pt]
\textbf{Trackers} & \textbf{Source}   & \textbf{AUC}  &\textbf{NPR}   &\textbf{PR} \\
\hline 
\textbf{CSWinTT}\cite{song2022transformertrackingcyclicshifting}    &IJCAI 2022           &66.2      &75.2     &70.9          \\  
\textbf{OSTrack}\cite{ye2022ostrack}    &ECCV 2022           &69.1       &78.7      &75.2         \\ 
\textbf{SeqTrack}\cite{chen2023seqtrack}    &CVPR 2023           &69.9       &79.7      &76.3         \\ 
\textbf{AQATrack}\cite{xie2024autoregressive}    &CVPR 2024           &71.4       &81.9      &78.6         \\ 
\textbf{TransVLT}\cite{zhao2023transformer}    &PRL 2023           &66.4       &-      &70.8         \\ 
\textbf{JointNLT}\cite{zhou2023joint}    &CVPR 2023           &60.4       &69.4      &63.6         \\ 
\textbf{MMTrack-384}\cite{zheng2023toward}    &TCSVT 2023           &70.0       &82.3      &75.7         \\ 
\textbf{ATTrack}\cite{ge2024consistencies}    &MM 2024           &63.7       &-      &67.3         \\ 
\textbf{OSDT}\cite{zhang2024one}    &TCSVT 2024           &64.3       &68.6      &73.4         \\ 
\textbf{UVLTrack-B}\cite{ma2024unifying}    &AAAI 2024           &69.4       &-      &74.9         \\ 
\textbf{UVLTrack-L}\cite{ma2024unifying}    &AAAI 2024           &71.3       &-      &78.3        \\ 
\textbf{QueryNLT}\cite{shao2024context}    &CVPR 2024           &59.9       &69.6      &63.5         \\ 
\textbf{DUTrack}\cite{Li2025dutrack}    &CVPR 2025           &73.0       &83.8      &81.1         \\ 
\textbf{ATSTrack}\cite{zhen2025atstrack}    &arxiv 2025           &72.6       &82.4      &79.5         \\ 
\textbf{GLAD }\cite{luo2026glad}   &IJCV 2026      &  69.5    & 79.6  & 74.2 \\ 
\textbf{ADTrack }\cite{zhang2026aware}   &AAAI 2026      & 71.5     & -  & 78.5 \\ 
\textbf{SDTrack }\cite{bai2026selective}   &ESWA 2026      &  71.5    &  - & 78.7 \\ 
\textbf{ABOTrack }\cite{hu2026vision}   &JVCIR 2026      &   69.5   & 79.9  & 76.7 \\ 
\textbf{SemaTrack }\cite{zhang2026sematrack}   &TVC 2026      &  -    & 77.4  & 80.3\\ 
\textbf{RrmTrack }\cite{zhang2026rwkv}   &TMM 2026      &   68.2   & -  & 74.5\\ 
\hline
\textbf{Ours }   & -      & 71.7  & 83.1   &  80.3 \\
\hline \toprule [0.5 pt]
\end{tabular}
\end{table}

$\bullet$ \textbf{Results on OTB99 Dataset.~} 
Table~\ref{OTB99_results} presents the comparison on OTB99-Lang~\cite{li2017tracking}. OTB99-Lang extends the classical OTB-style tracking evaluation with sentence-level language descriptions for 99 sequences, making it a compact but widely used benchmark for short- and medium-term language-guided tracking. Our method achieves 94.8 PR and 72.4 AUC. Although the PR is slightly lower than ReasoningTrack, our method obtains competitive AUC among recent trackers, exceeding ReasoningTrack by 1.3 points and DUTrack by 1.5 points. This result shows that the proposed text update mechanism can improve the overall overlap quality of tracking predictions while maintaining competitive precision.

\begin{table}
\center
\small   
\caption{Overall tracking performance on OTB99 dataset. } 
\label{OTB99_results}
\begin{tabular}{l|l|ccccc}
\hline \toprule [0.5 pt]
\textbf{Trackers} & \textbf{Source}   & \textbf{PR}   &\textbf{AUC} \\
\hline
\textbf{LSTMTrack}\cite{feng2020real}    &WACV 2020           &79.0       &61.0           \\ 
\textbf{SNLT}\cite{feng2021siamese}    &CVPR 2021           &84.8       &66.6            \\ 
\textbf{GTI}\cite{yang2020grounding}    &TCSVT 2021           &73.2       &57.1          \\ 
\textbf{TNL2k-II}\cite{wang2021tnl2k}    &CVPR 2021           &88.0       &68.0           \\ 
\textbf{TransVLT}\cite{zhao2023transformer}    &PRL 2023           &91.2       &69.9          \\ 
\textbf{JointNLT}\cite{zhou2023joint}    &CVPR 2023           &85.6       &65.3           \\ 
\textbf{MMTrack-384}\cite{zheng2023toward}    &TCSVT 2023          &91.8       &70.5            \\
\textbf{TDCL }\cite{wang2024tdcl}   &ECAI 2024      & 90.2  & 69.1     \\
\textbf{ATTrack}\cite{ge2024consistencies}    &MM 2024          &90.3       &69.3        \\ 
\textbf{OSDT}\cite{zhang2024one}    &TCSVT 2024          &86.7       &66.2            \\ 
\textbf{UVLTrack-B}\cite{ma2024unifying}    &AAAI 2024          &89.9       &69.3           \\ 
\textbf{QueryNLT}\cite{shao2024context}    &CVPR 2024          &88.2       &66.7            \\ 
\textbf{DUTrack}\cite{Li2025dutrack}    &CVPR 2025          &93.9       &70.9           \\ 
\textbf{ATSTrack}\cite{zhen2025atstrack}    &arXiv 2025          &94.4       &71.0           \\ 
\textbf{ReasoningTrack}\cite{wang2025reasoningtrack}   
& arXiv 2025      & 95.6  & 71.1            \\
\textbf{GLAD }\cite{luo2026glad}   
& IJCV 2026      & 92.4  & 69.6    
\\
\textbf{ADTrack }\cite{zhang2026aware}   
&AAAI 2026      & 93.9  & 72.3    
\\
\textbf{SDTrack }\cite{bai2026selective}   
&ESWA 2026      & 94.2  & 72.7    
\\
\textbf{ABOTrack }\cite{hu2026vision}   
&JVCIR 2026      & 91.1  & 71.2    
\\
\textbf{SemaTrack }\cite{zhang2026sematrack}   
&TVC 2026      & 94.5  & 72.2    
\\
\textbf{RrmTrack }\cite{zhang2026rwkv}   
&TMM 2026      & 93.7  & 71.2    
\\
\hline
\textbf{Ours }   & -      & 94.8  & 72.4    \\
\hline \toprule [0.5 pt]
\end{tabular}
\end{table}

\subsection{Ablation Study}

We conduct ablation studies on TNLLT to verify the contribution of each proposed component. The analysis focuses on four key aspects: the overall component effectiveness, the role of Qwen-based triplet refinement, the influence of target-conditioned sparse evidence selection, and the design choice of concept-span update.

\begin{figure*}
\centering
\includegraphics[width=\linewidth]{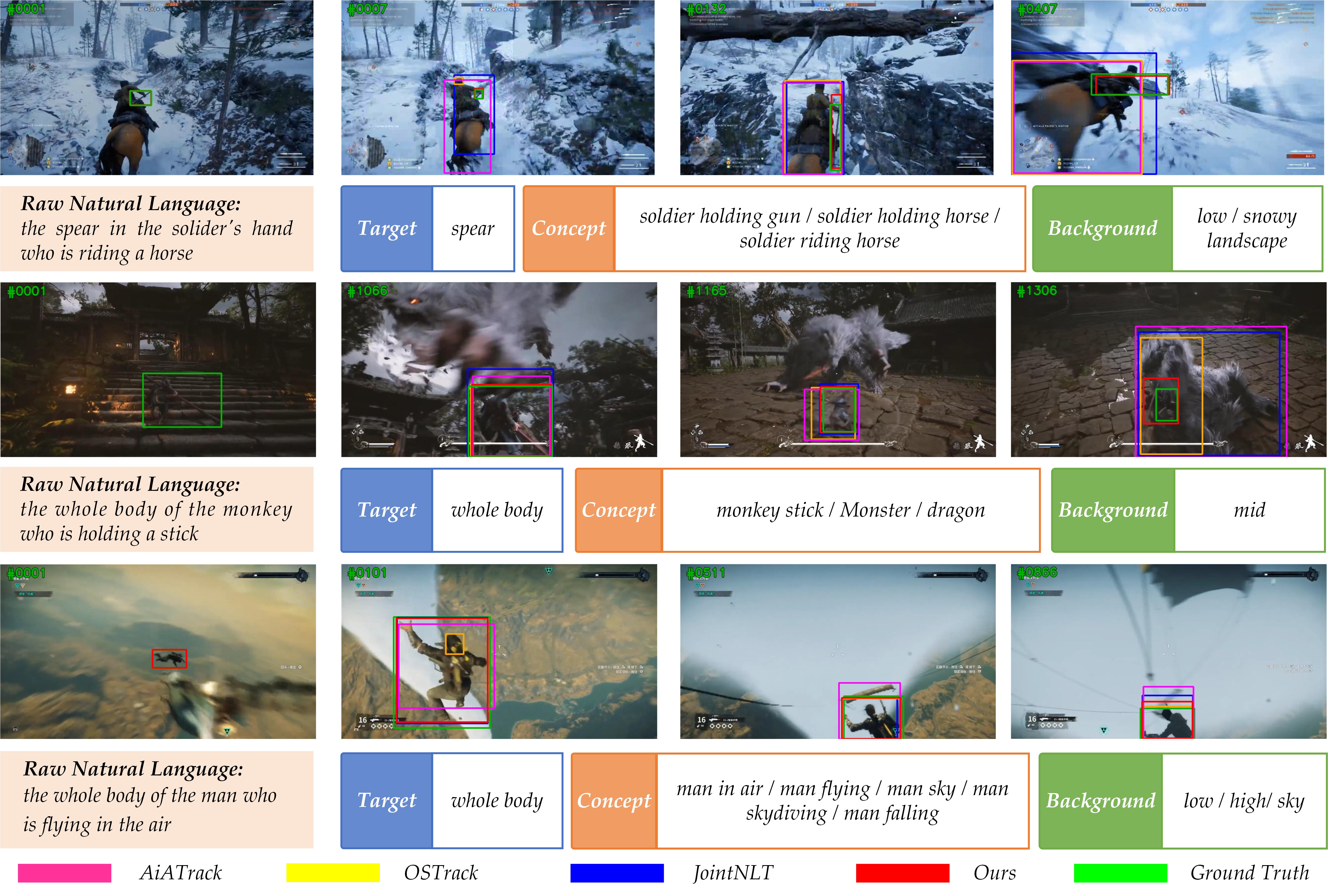}
\caption{\textbf{Qualitative comparison on TNLLT.} We compare our tracker with three representative trackers, including AiATrack, OSTrack, and JointNLT, on three challenging sequences. Each row corresponds to one video sequence and contains sampled keyframes, the raw language query, the structured triplet used by our method, and tracking results from different trackers. In the structured triplet, the target field captures stable object identity, the concept field describes appearance-related cues, and the background field provides contextual information. The visualized results show that our method localizes the target more accurately under challenging conditions such as motion blur, occlusion, scale variation, and background interference.}
\label{TrackingVis}
\end{figure*}

$\bullet$ \textbf{Component Analysis.~} 
Table~\ref{tab:ablation_component} reports the component-wise analysis of the proposed framework. Starting from the baseline tracker, introducing the DP-based structured triplet brings a consistent improvement from 73.9/76.5/63.3 to 74.1/77.3/63.9 in PR/NPR/AUC, showing that decomposing the raw query into role-specific fields provides more explicit language guidance than using the original sentence alone. When concept update is further added, the performance increases to 74.3 PR, 77.5 NPR, and 64.1 AUC, indicating that appearance-related concepts benefit from visual-conditioned refinement. By adding target-conditioned sparse Top-K evidence selection, the tracker reaches 74.8 PR, 78.0 NPR, and 64.4 AUC. This demonstrates that using the target field as a stable identity anchor helps select more relevant visual tokens for concept updating. Finally, incorporating Qwen refinement leads to the best performance, achieving 75.0 PR, 78.2 NPR, and 64.5 AUC. These progressive improvements verify that the proposed modules are complementary and jointly contribute to more robust language-guided tracking.

\begin{table}[t]
\centering
\caption{Component analysis of the proposed framework on TNLLT.}
\label{tab:ablation_component}
\resizebox{\linewidth}{!}{%
\begin{tabular}{cccccc|ccc}
\hline \toprule [0.5 pt]
\#No. & DP & Qwen & C-Upd & T-Cond & Top-K & PR & NPR & AUC \\
\midrule
01 & & & & & & 73.9 & 76.5 & 63.3 \\
02 & \checkmark & & & & & 74.1 & 77.3 & 63.9 \\
03 & \checkmark & & \checkmark & & & 74.3 & 77.5 & 64.1 \\
04 &\checkmark & & \checkmark & \checkmark & \checkmark & 74.8 & 78.0 & 64.4 \\
05 & \checkmark & \checkmark & \checkmark & \checkmark & \checkmark & 75.0 & 78.2 & 64.5 \\
\hline \toprule [0.5 pt]
\end{tabular}%
}
\end{table}



$\bullet$ \textbf{Effect of Qwen-based Triplet Refinement.~} 
Table~\ref{tab:ablation_qwen} studies different Qwen refinement strategies. Directly using a frozen Qwen refiner decreases the performance to 72.8 PR, 76.06 NPR, and 62.69 AUC, which suggests that general-purpose language rewriting is not necessarily aligned with the tracking objective and may introduce visually irrelevant or semantically unstable descriptions. In contrast, tracking-aware LoRA fine-tuning significantly improves the refinement quality. Among different LoRA ranks, the rank-16 variant achieves the best result with 75.04 PR, 78.16 NPR, and 64.51 AUC, outperforming both rank-8 and rank-32 settings. This indicates that a moderate adaptation capacity is sufficient to learn tracking-oriented triplet refinement, while overly weak or overly large adapters may lead to under-adaptation or less stable refinement behavior.

\begin{table}[t]
\centering
\caption{Effect of Qwen-based triplet refinement on TNLLT.}
\label{tab:ablation_qwen}
\resizebox{\linewidth}{!}{
\begin{tabular}{p{5.2cm}lccc}
\hline \toprule [0.5 pt]
Setting & PR & NPR & AUC \\
\midrule
No Qwen refinement  & 74.8 & 78.0 &
64.4 \\
Frozen Qwen & 72.8 & 76.06 & 62.69 \\
Qwen-LoRA (r=8) & 73.8 & 77.13 & 63.66 \\
Qwen-LoRA (r=16) & 75.04 & 78.16 & 64.51 \\
Qwen-LoRA (r=32) & 74.34 & 77.59 & 64.06 \\
\hline \toprule [0.5 pt]
\end{tabular}
}
\end{table}

$\bullet$ \textbf{Effect of Target-Conditioned Top-K Evidence Selection.~}
The Top-K evidence selection module controls how much search-region visual evidence is written back to the concept field. As shown in Table~\ref{tab:ablation_topk}, the dense update performs the worst, obtaining only 71.9 PR, 74.9 NPR, and 61.8 AUC. This indicates that using all search tokens introduces substantial background noise and distractor information into the language update. Sparse target-conditioned selection consistently improves the results. With $K=4$, the tracker reaches 74.1 PR, 77.2 NPR, and 63.8 AUC, showing that a small set of target-relevant tokens is already more effective than dense writeback. Increasing $K$ to 8 gives the best performance, achieving 74.8 PR, 78.0 NPR, and 64.4 AUC. Further increasing the evidence set to $K=12$ and $K=16$ slightly degrades the results, suggesting that overly large token sets start to reintroduce irrelevant visual context. Therefore, we adopt $K=8$ as the default setting, which provides a balanced trade-off between sufficient target evidence and controlled language update noise.

\begin{table}[t]
\centering
\caption{Effect of target-conditioned Top-K evidence selection on TNLLT.}
\label{tab:ablation_topk}
\resizebox{\linewidth}{!}{
\begin{tabular}{p{5.2cm}lccc}
\hline \toprule [0.5 pt]
Setting & PR & NPR & AUC \\
\midrule
ST + target-conditioned dense update & 71.9 & 74.9 & 61.8 \\
ST + target-conditioned Top-4 update & 74.1 & 77.2 & 63.8 \\
ST + target-conditioned Top-8 update & 74.8 & 78.0 & 64.4 \\
ST + target-conditioned Top-12 update & 74.5 & 77.6 & 64.2 \\
ST + target-conditioned Top-16 update & 74.0 & 77.0 & 63.7 \\
\hline \toprule [0.5 pt]
\end{tabular}
}
\end{table}

$\bullet$ \textbf{Effect of Concept Span Update Design.~} 
Table~\ref{tab:ablation_update_design} further analyzes how the structured fields should be updated. Here, ST denotes structured triplet parsing, C and B denote concept and background fields, Span denotes span-local writeback, and T-Top8 denotes target-conditioned Top-8 evidence selection. Using the structured triplet alone improves the baseline, but directly updating both concept and background fields leads to a clear performance drop. This is because the background field mainly describes spatial context and scene layout, which is less stable under long-term tracking and can introduce distracting context when updated aggressively. Applying span-local writeback alleviates this issue, but the joint concept-background update is still inferior to concept-only update. The best non-Qwen setting is obtained by updating only the concept field with target-conditioned Top-8 evidence, reaching 74.8 PR, 78.0 NPR, and 64.4 AUC. These results support our design choice: the target field should remain stable, the concept field should be visually updated, and the background field should be used as auxiliary context rather than an actively updated appearance description.

\begin{table}[t]
\centering
\caption{Effect of concept update design on TNLLT.}
\label{tab:ablation_update_design}
\resizebox{\linewidth}{!}{
\begin{tabular}{
  p{\dimexpr 5.2cm/5 - 2\tabcolsep\relax}
  p{\dimexpr 5.2cm/5 - 2\tabcolsep\relax}
  p{\dimexpr 5.2cm/5 - 2\tabcolsep\relax}
  p{\dimexpr 5.2cm/5 - 2\tabcolsep\relax}
  p{\dimexpr 5.2cm/5 - 1\tabcolsep\relax}
|ccc}
\hline \toprule [0.5 pt]
ST & C & B & Span & T-Top8 & PR & NPR & AUC \\
\midrule
&  &  &  &  & 73.9 & 76.5 & 63.3 \\
\checkmark &  &  &  &  & 74.1 & 77.3 & 63.9 \\
\checkmark & \checkmark & \checkmark &  &  & 72.4 & 75.5 & 62.1 \\
\checkmark & \checkmark & \checkmark & \checkmark &  & 73.5 & 76.7 & 63.4 \\
\checkmark & \checkmark &  & \checkmark &  & 74.3 & 77.5 & 64.1 \\
\checkmark & \checkmark &  & \checkmark & \checkmark & 74.8 & 78.0 & 64.4 \\
\hline \toprule [0.5 pt]
\end{tabular}
}
\end{table}



\begin{figure}[t]
\centering
\includegraphics[width=\linewidth]{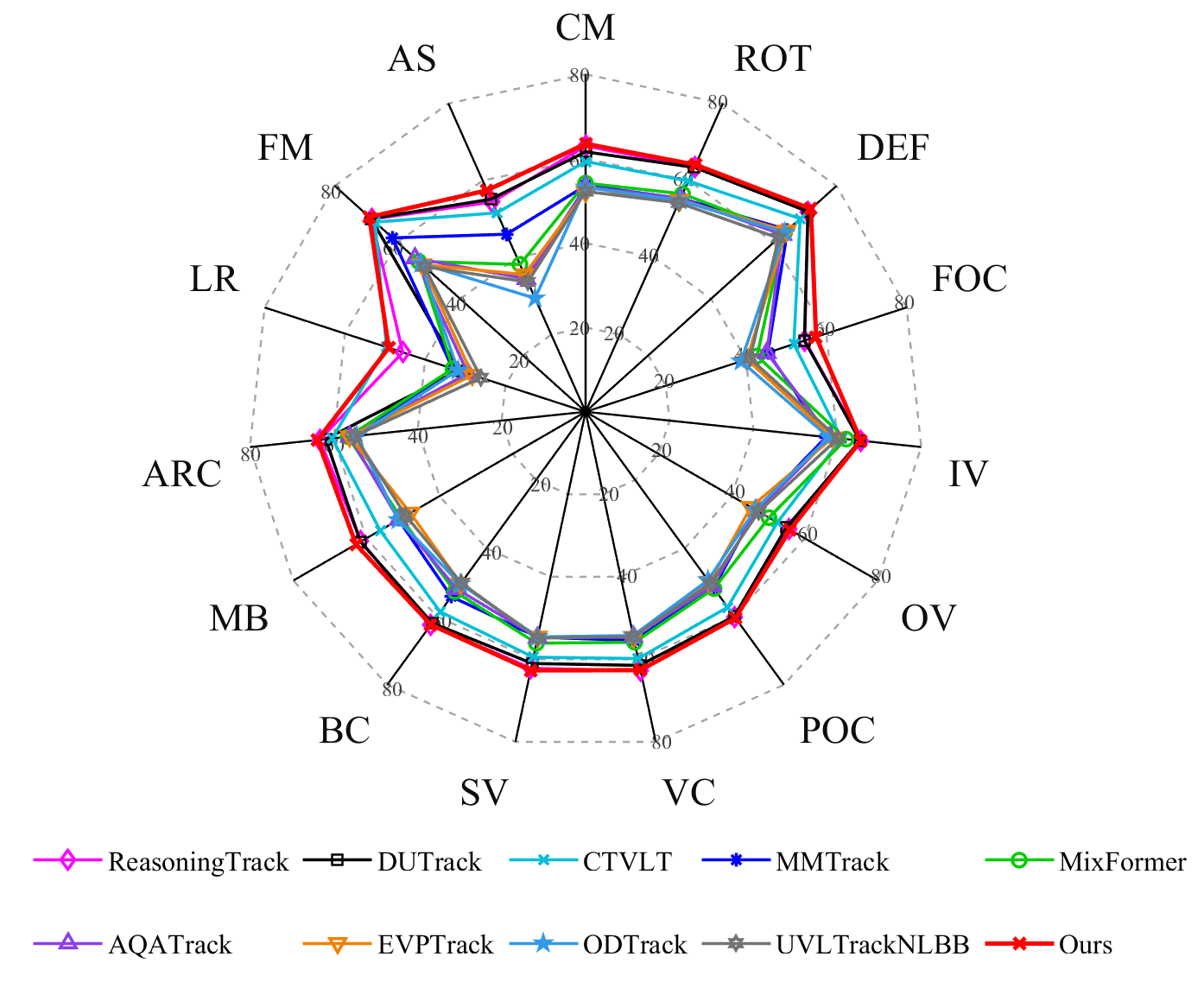}
\caption{\textbf{Attribute-based robustness analysis on TNLLT.} The radar chart compares different trackers across 15 challenging attributes. Larger values indicate better robustness under the corresponding attribute.}
\label{fig:attribute_radar}
\end{figure}

$\bullet$ \textbf{Attribute-based Robustness Analysis.~} 
To further analyze the robustness of different trackers under diverse long-term tracking challenges, we report the attribute-level performance of representative trackers on 15 attributes of the TNLLT dataset. These attributes include CM, ROT, DEF, FOC, IV, OV, POC, VC, SV, BC, MB, ARC, LR, FM, and AS, covering various challenging factors such as appearance variation, occlusion, motion disturbance, scale change, and background interference. As shown in Fig.~\ref{fig:attribute_radar}, our method is compared with representative trackers, including ReasoningTrack, DUTrack, CTVLT, MMTrack, MixFormer, AQATrack, EVPTrack, ODTrack, and UVLTrackNLBB, in terms of attribute-level performance.

Our method achieves consistent improvements on most attributes and obtains the best results on 12 out of 15 attributes. Specifically, it performs well under appearance and geometric variations, achieving scores of 63.52 on CM, 64.10 on ROT, 71.88 on DEF, and 65.64 on IV. For occlusion-related challenges, our method also achieves favorable results, with scores of 57.44 on FOC, 55.95 on OV, and 60.45 on POC. In addition, it shows clear advantages under motion-related attributes, achieving 62.85 on MB and 68.99 on FM. Although our method is slightly behind the best-performing tracker on VC, BC, and LR, the performance gaps are marginal. These results indicate that the proposed fine-grained language update strategy improves tracking robustness across diverse challenging conditions, rather than only improving the overall average performance.

\subsection{Visualization}

\noindent $\bullet$ \textbf{Language Description Parsing.~}  
Figure~\ref{LDP} provides intuitive examples of the proposed dependency-based language parsing process. For each raw query, the parser first builds token-level dependency relations and then assigns different linguistic components to role-specific fields. In the first example, the noun ``spear'' is selected as the target identity, while the phrase related to ``soldier's hand who is riding horse'' is initially grouped as the concept field. After refinement, the concept is converted into a more tracking-oriented description, i.e., ``soldier holding gun'', and the background field is supplemented with scene context such as ``snowy landscape''. Similar parsing and refinement behavior can be observed in the other examples, where the target field keeps stable object identity, the concept field emphasizes appearance or interaction cues, and the background field provides auxiliary spatial or scene information. These visualizations show that our parser does not simply split the sentence by word order, but uses syntactic relations to preserve the functional role of each phrase before constructing and refining the structured triplet.

\noindent $\bullet$ \textbf{Tracking Results.~}  
Figure~\ref{TrackingVis} visualizes tracking results on three challenging TNLLT sequences. Each row shows several representative frames together with the raw language query and the structured triplet produced by our language parsing and refinement pipeline. The target field describes the stable object identity, the concept field summarizes appearance-related cues, and the background field provides contextual information such as scene layout or relative position. Compared with existing trackers, our method produces more accurate bounding boxes under fast motion, low contrast, occlusion, and cluttered backgrounds. These qualitative results indicate that fine-grained language decomposition helps the tracker focus on the correct target while using concept and background information to suppress distractors.

\subsection{Limitation Analysis} 

Although the proposed framework achieves strong performance by introducing fine-grained text updating into vision-language tracking, several aspects can be further explored. First, the Qwen-based refinement module introduces additional language processing cost compared with directly using the original query. Developing more lightweight refinement models or more efficient updating strategies would further improve its deployment efficiency. Second, in extremely long-term scenarios where the target disappears for a long duration or visual evidence becomes highly unreliable, the tracker may still require stronger temporal reasoning and memory mechanisms to recover the target after reappearance. Third, our current framework mainly updates the language guidance according to the current search-region evidence. Incorporating richer historical observations and long-range temporal cues may further enhance the stability of language-guided tracking under complex target state transitions.

\section{Conclusion} 
This paper presents a fine-grained text update guided framework for long-term vision-language tracking. By decomposing the initial query into target, concept, and background components, our method provides a structured language state for more controllable tracking-oriented refinement. A Qwen-based refinement module improves the semantic quality of the parsed triplet, and a target-conditioned Top-K visual evidence selection mechanism further updates concept descriptions according to current visual observations while preserving stable target identity. Extensive experiments on multiple benchmarks show that the proposed framework consistently improves tracking performance, and ablation studies validate the effectiveness of each component. These results demonstrate the importance of fine-grained language decomposition and adaptive textual updating for robust vision-language tracking.

\section*{Acknowledgment} 
This work was supported by the National Natural Science Foundation of China under Grant 62102205. Anhui Provincial Natural Science Foundation-Outstanding Youth Project, 2408085Y032. The authors acknowledge the High-performance Computing Platform of Anhui University for providing computing resources.

\small{ 
\bibliographystyle{IEEEtran}
\bibliography{reference}
}

\end{document}